\documentclass[sn-mathphys]{sn-jnl}

\usepackage{graphics}
\usepackage{subfigure}
\usepackage{algorithm}
\usepackage{algorithmicx}
\usepackage{algpseudocode}

\jyear{2021}%

\theoremstyle{thmstyleone}%
\theoremstyle{thmstyletwo}%

\theoremstyle{thmstylethree}%

\begin{document}

\title[A Multi-Layer Regression based Predicable Function Fitting Network]{A Multi-Layer Regression based Predicable Function Fitting Network}


\author*[1]{\fnm{Changlin} \sur{Wan}}\email{wancl@hzu.edu.cn;wancl@21cn.com}

\author[2]{\fnm{Zhongzhi} \sur{Shi}}

\affil*[1]{\orgname{Huizhou University}, \orgaddress{\postcode{516007}, \state{Guangdong}, \country{China}}}

\affil[2]{\orgname{CAS Key Lab. of IIP, Institute of Computing Technology}, \orgaddress{\postcode{100190}, \state{Beijing}, \country{China}}}


\abstract{Function plays an important role in mathematics and many science branches. As the fast development of computer technology, more and more study on computational function analysis, e.g., Fast Fourier Transform, Wavelet Transform, Curve Function, are presented in these years. However, there are two main problems in these approaches: 1) hard to handle the complex functions of stationary and non-stationary, periodic and non-periodic, high order and low order; 2) hard to generalize the fitting functions from training data to test data. In this paper, a multiple regression based function fitting network that solves the two main problems is introduced as a predicable function fitting technique. This technique constructs the network includes three main parts: 1) the stationary transform layer, 2) the feature encoding layers, and 3) the fine tuning regression layer. The stationary transform layer recognizes the order of input function data, and transforms non-stationary function to stationary function. The feature encoding layers encode the raw input sequential data to a novel linear regression feature that can capture both the structural and the temporal characters of the sequential data. The fine tuning regression layer then fits the features to the target ahead values. The fitting network with the linear regression feature layers and a non-linear regression layer come up with high quality fitting results and generalizable predictions. The experiments of both mathematic function examples and the real word function examples verifies the efficiency of the proposed technique.}

\keywords{Function Regression, Linear Regression Feature, Stationary Transform, Predicable Function Fitting}



\maketitle

\section{Introduction}\label{sec1}
Function plays an important role in mathematics and many science branches \cite{RUTKOWSKI198533} \cite{DAVIS2021107083}. The computational function analysis, e.g., Fourier Transform, Wavelet Transform, Function Curve Fitting, and Function Fitting Neural Network are important techniques in mathematics and many science partitions. These approaches can both be used to solve the function fitting problems and many other application problems.
As we known that the Fast Fourier Transform (FFT) is widely applied in signal processing area \cite{Cooley1965AnAF}\cite{Brooks2012}, while the signals can be regarded as the data sets of time dependent function $s=f(t)$ or position dependent function $s=f(x,y,z)$. Since the Fourier transform decomposes the functions to a series of cosine/since basis functions, it is particularly suitable for the analysis of stationary or periodic functions but not for the non-stationary functions. Furthermore, the FFT did not take account into the general fitting error, it is usually has small fitting error in training data set but has large fitting error in test data set. There are still some works undone to apply it for generalizing the FFT from training data to test data. The Wavelet Transform (WT), which is invented as a seismic wave analysis and predict tool, can not only fit periodic functions but also non-periodic functions, especially suitable to the signals with sparse pulse \cite{Mallat1989}\cite{Yang:18}. However, it cannot handle non-stationary function, such as polynomial functions and exponential functions. The Curve Fitting (CF) method is another widely adopted function fitting method for its certainty and efficiency \cite{IGNACIO2021125681, Moghadam2019, Bayard94, pmlr-v162-d-ascoli22a}. It first defines the general form of fitting function, e.g., polynomial functions, trigonometric functions, or exponent functions, and then use a parameter optimizer to find the optimized parameters of the predefined fitting functions to minimize the fitting error. This approach usually has better generalizing performance if it has the correct function forms. However, the form of functions is usually a posterior-knowledge that is unavailable and a limit to fitting error. The recent deep neural network is a powerful function fitting theory\cite{hochreiter1997long, KaiserB16, LaiCYL18, ZhouZPZLXZ21, WuXWL21}, it achieves the advanced results on many function fitting applications, such as image classification, text analyze, and time series. However, there are still few study of deep neural network on function analysis, especially on those high order complex functions.

It can be concluded that there are two main problems in these approaches: 1) hard to handle the both functions of stationary and non-stationary\cite{SharmanF84} \cite{Hang2022}; 2) hard to generalize the fitting functions from training data to test data\cite{GaoDYS21}. In this paper, a linear regression feature based approach that solves the two main problems is introduced as a predicable function fitting and analyzing technique. The rest of the paper is organized as follows: Section \ref{sec-2} try to give a formal definition of the predicable function problem; Section \ref{sec-3} describes the theorems and methods of the linear regression feature based predicable function fitting technique; Section \ref{sec-4} provides and analysis the experiment results of the presented method; Section \ref{sec-5} makes the conclusion and discussion.

\section{Predicable Function Fitting Problem Definitions}\label{sec-2}
The predicable function fitting problem can be defined as follows:\\
given a mathematics function $y=f(x)$, $x$ is a sort-able series variable, try to approximate or fit the function by a new fitting function $y=f^*(x)$ from a training data set of $x-y$ pairs $(x_i,y_i=f(x_i)),i=1,...,n$. The new fitting function $y=f^*(x)$ can approximate or generalize the function $y=f(x)$ to the test data set of $x-y$ pairs $(x_j,y_j=f(x_j)),j=n+1,...,n+m$, such that the general fitting error $E=\sum_{j=n+1}^{n+m}(f(x_j)-f^*(x_j))^2$ is minimized.\\

As shown in the problem definition, the new fitting function is not only needed to reconstruct the known training data of target function, but also needed to have the ability to predict or generalize to the unknown test data. This is guaranteed by the minimization of the general fitting error enforced on the test data.

To solve this problem, a function analysis model that can define the form of the new functions, and a parameter regressor that can define the parameters of the new functions are both needed. In the literature, the function analysis models include: Fourier Transform (FT), Wavelet Transform (WT), Curve Fitting (CF), and Neural Network (NN); and the parameter regressor include: Linear Regressor, Boosted Regressor and so on. For the regressor, the general fitting error is often approximated by the training fitting error with regularization technology to prevent over-fitting.

\section{Linear Regression Transform based Predicable Function Fitting}\label{sec-3}
\subsection{Linear Regression Feature}
For the predicable function fitting problem, the fitting regressor needs the feature input data to compute the output prediction. As far as we know, the raw training sequential $(x,y)$ pairs are usually not sufficient. However, there is no efficient feature schema except those embedded features of the sequential analysing neural network, fore example Long-Short-Term-Memory network. Those neural networks can learn features through the input raw sequential data in a iterative and black-box way. The main problems to design such a feature schema is that the feature should not only have the memory of the history data but also can represent the positional structure of the data. In this section, a linear Regression feature schema that addresses this main problem is proposed. In this schema, the feature includes multiple dimensions. At a position on the input sequential, each dimension of the feature vector is a linear regression result of the past sequential data and the past feature dimensions. Given a function $y=f(x)$ with sequential input $x=1,2,...,n$, and the differential function $dy(x)=y(x)-y(x-1)$, the $m$-dimensions linear Regression feature $F(x,y)$ can be defined as:
\[
\begin{array}{rll}
F_{i}(x,y) & = & (v_{i1},v_{i2},...,v_{im}),\\
v_{i1} & = & R_1(y_{(i-m+1)},...,y_{i},dy_{(i-m+1)},...,dy_{i}),\\
v_{ij} & = & R_j(y_{(i-m-j+2)},...,y_{(i-j+1)},dy_{(i-m-j+2)},...,\\
        & & dy_{(i-j+1)},v_{i1},...,v_{ik}),j>1,k<j.\\
\end{array}
\]
where $R_k, k=1,...,m$ is a linear regression function defined as:
\[
\begin{array}{rll}
R_k & = & a_{1}\times y_{i}+...+a_{m}\times y_{(i-m+1)}\\
  & & + b_{1}\times dy_{i}+...+b_{m}\times dy_{(i-m+j)}\\
  & & + c_{1}\times v_{i1}+...+c_{k}v_{ik}+ d\\
  & & == y_{(i+1)}.
\end{array}
\]

An illustrative example of 4-dimensions feature computation procedure is shown in the following figure.

\begin{figure}[h]%
\centering
\includegraphics[width=0.9\textwidth]{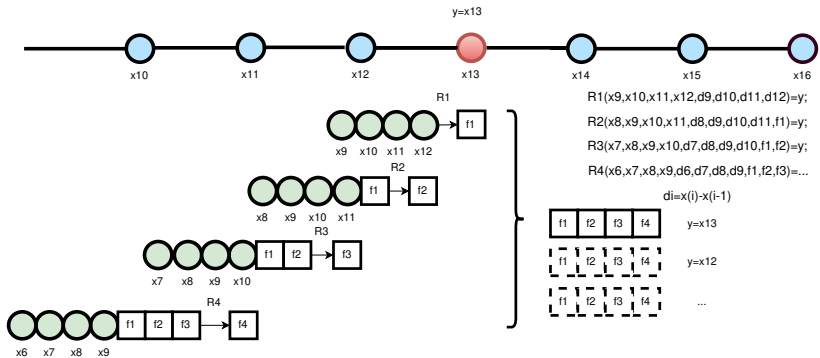}
\caption{illustrative example of 4-dimensions feature computation procedure}\label{fig1}
\end{figure}

\subsection{Train the Function Fitting Regressor}\label{sec-32}
The linear Regression feature procedure decomposing the function $y=f(x)$ to a set of linear functions $R_k$. Each $R_k$ is a linear function trained by a linear regression procedure with backward data as training input and forward data as target output. The function $R_k$ can approximate the target value through the linear transform, and the transform result is used as a dimension of linear Regression feature $F_k$. Thus the linear Regression transform procedure transforms the training x-y pairs to a set of feature vectors $t(x_i)=(F_1(i),F_2(i),...)$. These vectors are used as the input feature data of the function fitting regressor in one hand. In the other hand, the target values of the regressor are computed from the one-step-ahead value of the given function. With the input feature data $F=(F_1,F_2,...)$ and input target values $Y$, a function fitting regressor $R$ can be trained by a regression optimizer, such as function fitting neural network (fitnet) regressor \cite{BassonE99}, LSBoost regressor \cite{Breiman01}, and XGBoost regressor \cite{chen2016}. These regression optimizer served as a non-linear fine-tune layer, which can fitting the non-linear functions based on the linear regression features.

\subsection{New Data Generalization}\label{sec-33}
Given a function within a sliding window, the aim of new data generalization is to compute the in the future range of using the new fitting function. Since we have obtained the trained regressor as the fitting function, the new data generalization algorithm can be described as follows:

\begin{algorithm}
\caption{(New Data Generalization)}\label{algo2}
\begin{algorithmic}[2]
\Require $y=f(x),x\in [x_1,x_2,...,x_n]$; regressor $R$; steps $m$
\Ensure $y=f(x),x\in [x_1,x_2,...,x_{n+m}]$
\State set the initial range partition $X = [x_i,x_2,...,x_{n+m}]$;
\State set the initial estimate function $y = f(x), x\in X$;
\For{$i=0:m-1$}
    \State move the sliding window to $[x_{i+1},x_{i+2},...,x_{i+n}]$;
    \State apply the linear Regression transform to get the transformed feature vector $t(x_{i+n})$ of $x_{i+n}$;
    \State apply the regressor $R$ on $t(x_{i+n})$ the get the prediction $p_i$;
    \State use the prediction $p_i$ to compute the next value $y_{i+n+1}$;
    \State update $(x_{i+n+1},y_{i+n+1})$ on the function $y=f(x)$;
\EndFor

\end{algorithmic}
\end{algorithm}

\subsection{Stationary Transform}\label{sec-34}
Since the current regression optimizer is stationary process, it can usually map a finite range of feature data to a finite range of target values. To the end of predicable function fitting, the fitting function should be able to approximate a wide range of $y=f(x)$, e.g, $O(x^{-3})$, $O(x^{10}$, $O(e^x)$. To cope with this situation, a stationary transform is developed to map the functions with different orders into the nearly stationary functions. It uses a set of $log$ operators, $exp$ operators and linear operators to approximate a function with different orders with a nearly stationary function. As the pre-processing before the Linear Regression Decomposition, this stationary transform procedure provides the nearly stationary functions $S(x)$ to the Linear Regression Decomposition procedure. Given a function $y=f(x)$ within a sliding window $\left [x_1,x_2,...,x_n\right ]$, the Stationary Transform procedure is described as follows:

\begin{algorithm}
\caption{(Stationary Transform)}\label{algo3}
\begin{algorithmic}[3]
\Require $y=f(x),x\in [x_1,x_2,...,x_n]$;
\Ensure stationary transform function $y=S(x)$;
\State set the initial range partition $X = [x_i,x_2,...,x_{n+m}]$;
\State set the initial estimate function $y = f(x), x\in X$;
\For{$i=0:m-1$}
    \State compute $Y1 = f(x)+1$;
    \State estimate the polynomial order $k$ from $y=f(x)$;
    \State use $x^k$ to fit $Y1$;
    \State estimate the exponential base $b$ from $y=f(x)$;
    \State use $b^x$ to fit $Y1$;
    \If{the fit error of $x^k$ is less than the fit error of $b^x$}
        \State fit $x1=x,y1=log(diff(Y1))$ by a linear function $y1=a*x1+b1$;
        \State fit $x2=cumsum(exp(y1)),y2=Y1$ by a linear function $y2=a2*x2+b2$;
        \State $S(x)=\frac{f(x)+1}{a2*cumsum(exp(a1*x+b1))+b2}$;
    \Else
        \State fit $x1=x,y1=log(Y1)$ by a linear function $y1=a*x1+b1$;
        \State fit $x2=exp(y1),y2=Y1$ by a linear function $y2=a2*x2+b2$;
        \State $S(x)=\frac{f(x)+1}{a2*(exp(a1*x+b1)+b2}$;
    \EndIf
\EndFor

\end{algorithmic}
\end{algorithm}
It is notable that the variables sometimes not given, in this situation, a natural number sequence of $x=[1,2,...,n]$ is set as the default training range of $x$, a sequence of $x=[1,2,...,n+m]$ is set the default test range of $x$. Additionally, both the time and the space cost of this algorithm are $O(N)$.

\subsection{Predicable Function Fitting Framework}\label{sec-35}
\begin{figure}[h]%
\centering
\includegraphics[width=0.9\textwidth]{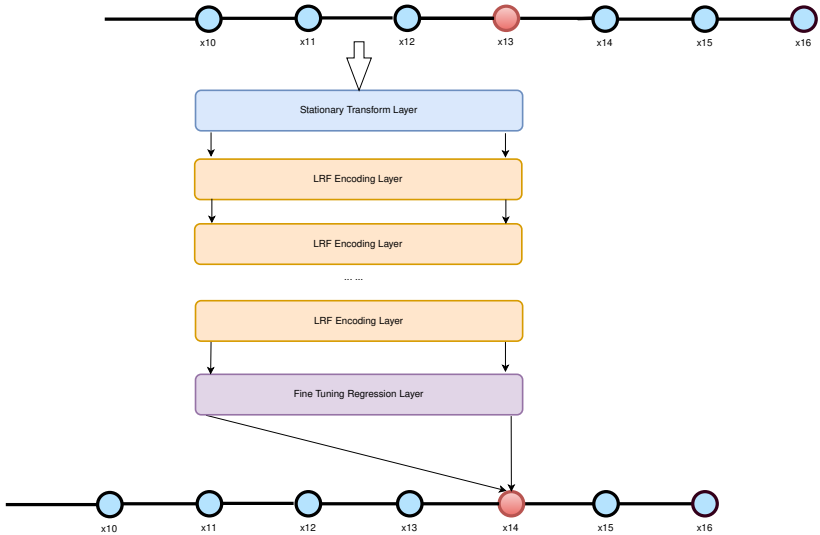}
\caption{Predicable Function Fitting Network Architecture}\label{fig2}
\end{figure}

Having the above procedures, the network architecture can be illustrated as Fig. \ref{fig2}.
As shown in Fig. \ref{fig2}, the network includes three main parts: 1) the stationary transform layer, 2) the feature encoding layers, and 3) the fine tuning regression layer. The stationary transform layer recognizes the order of input function data, and transforms non-stationary function to stationary function. The feature encoding layers encode the raw input sequential data to a novel linear regression feature that can capture both the structural and the temporal characters of the sequential data. The fine tuning regression layer then fits the features to the target ahead values.

In addition, the framework of the linear Regression transform and Regression based predicable function fitting can be illustrated as Fig. \ref{fig3}.

\begin{figure}[h]%
\centering
\includegraphics[width=0.9\textwidth]{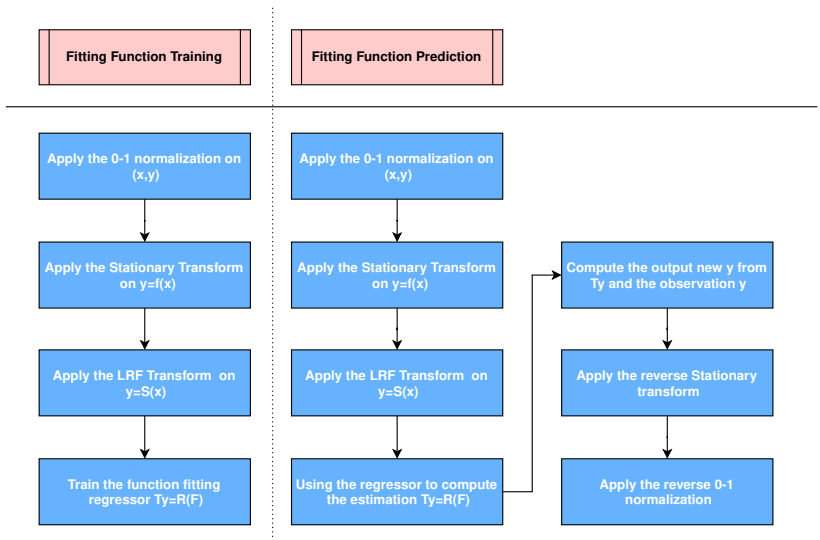}
\caption{Predicable Function Fitting Framework}\label{fig3}
\end{figure}

As shown in Fig. \ref{fig3}, the framework is formed by two parts, the one is fitting function training part, and the other is fitting function generalization part. The aim of training part is to train a regressor to approximate the given functions by LRF based function decomposition. Before the actual training procedure, the 0-1 normalization, stationary transform, LRF transform procedures are applied. Considering the time and space complexity of the ST transform, LRF transform and generalization regressor, they all can be performed by a Gradient Descent optimization algorithm of complexity $O(N)$. The aim of generalization part is to compute the new function value form the past function value by a LRF based function reconstruction. This part includes the 0-1 normalization, stationary transform, LRF transform, and computation, reverse stationary transform, reverse 0-1 normalization.

\section{Computational Experiments}\label{sec-4}
To evaluate the correctness and robustness of the proposed approach, some mathematics functions fitting and real world functions fitting computational experiments are conducted. The mathematics functions include:$f_1:y=x^3+3x^2-10x$, $f_2:y=x^{10}$,  $f_3:y=x^{-3}$, $f_4:y=e^{x}$, $f_5:y=sin(0.001x\pi+0.5\pi)+2cos(0.002x\pi+0.1\pi)$, $f_6:y=sin(0.001x\pi+0.5\pi)+2cos(0.002x\pi+0.1\pi+0.001x\pi)$). For $f_1,f_2,f_3,f_4$, the training range of $x$ is $[1:0.01:10]$, and the test range of $x$ is $[1:0.01:20]$. For $f_5,f_5$, the training range of $x$ is $[0:3050]$, and the test range of $x$ is $[0:4050]$. The single column data of these data set are treated as variable $y$, the data index $[1:N]$, $N$ is the number of data items,  are treated as variable $x$. The proposed approach is used to compute the new value $y$ of the test data $[1:N+1000]$.

The real-world functions include 8 time-series prediction data sets: the data set of the international airline passengers, the data set of 240 years of solar spots, the ETT \cite{ZhouZPZLXZ21} dataset contains the electricity transformers in every 15 minutes between July 2016 and July 2018, the Electricity1 dataset contains the hourly electricity consumption from 2012 to 2014, the Exchange \cite{LaiCYL18} dataset includes the daily exchange rates from 1990 to 2016, the Traffic data set includes the hourly data from California Department of Transportation, the Weather data set includes every 10 minutes temperature for 2020 year, the ILI data set includes the weekly recorded influenza-like illness (ILI) patients data from the United States between 2002 and 2021.

For the sake of fairness, the LRF based method uses three state-of-art regression optimizers for the tests. The one regressor is the function fitting neural network `fitnet', the two is ensemble regression method ‘LSBoost’, the three is the popular boost regression method `XGBoost'. The proposed approach performs the predicable function fitting experiments on each of the functions and each of the regressor. The LSTM neural network and the Informer Transformer network are used as the baseline. The experiment results are shown as follows:
\begin{figure*}[!t]
\centering
\subfigure[f1]{\includegraphics[width=.3\linewidth]{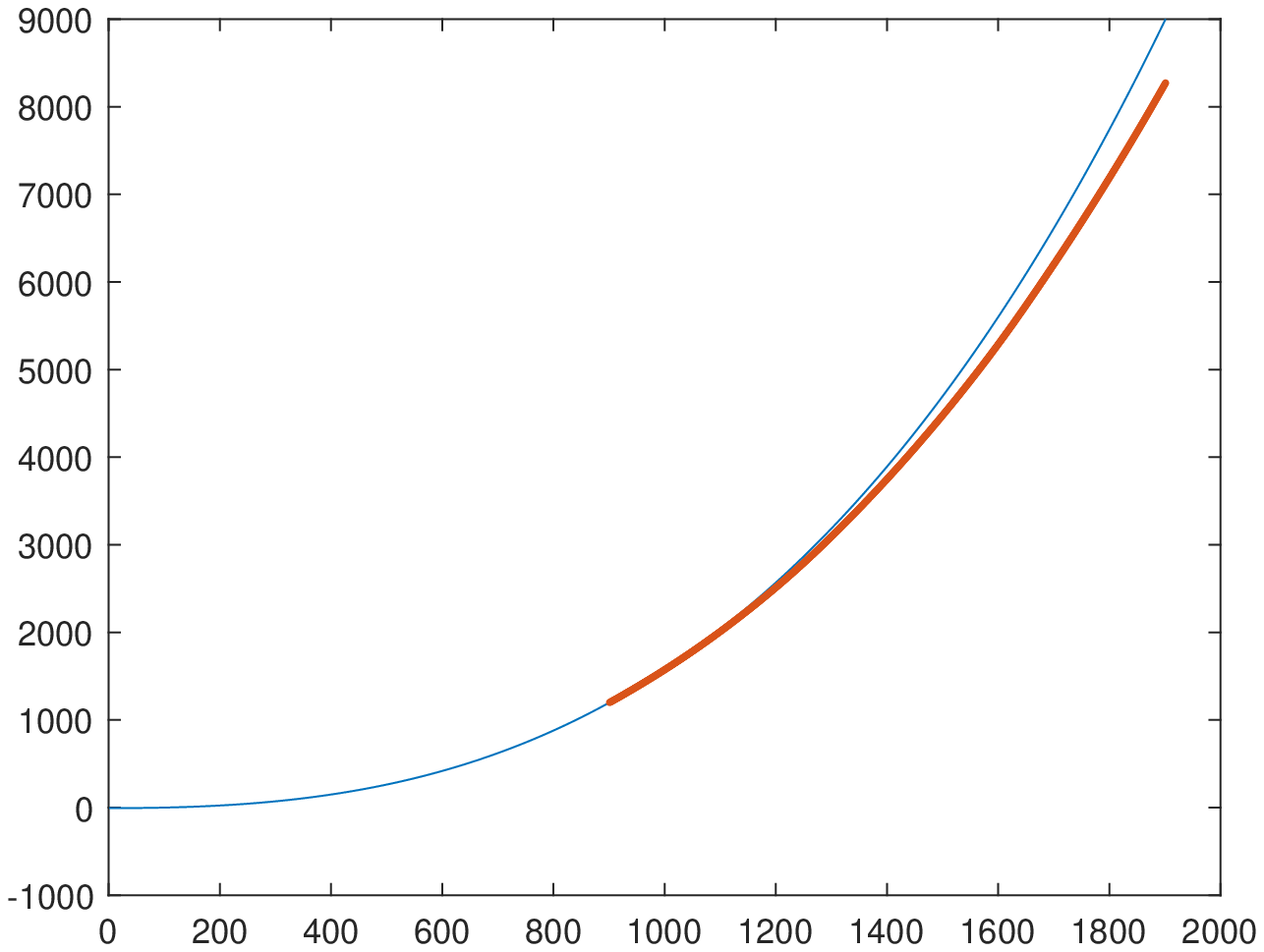}%
\label{fig3:f1}}
\subfigure[f2]{\includegraphics[width=.3\linewidth]{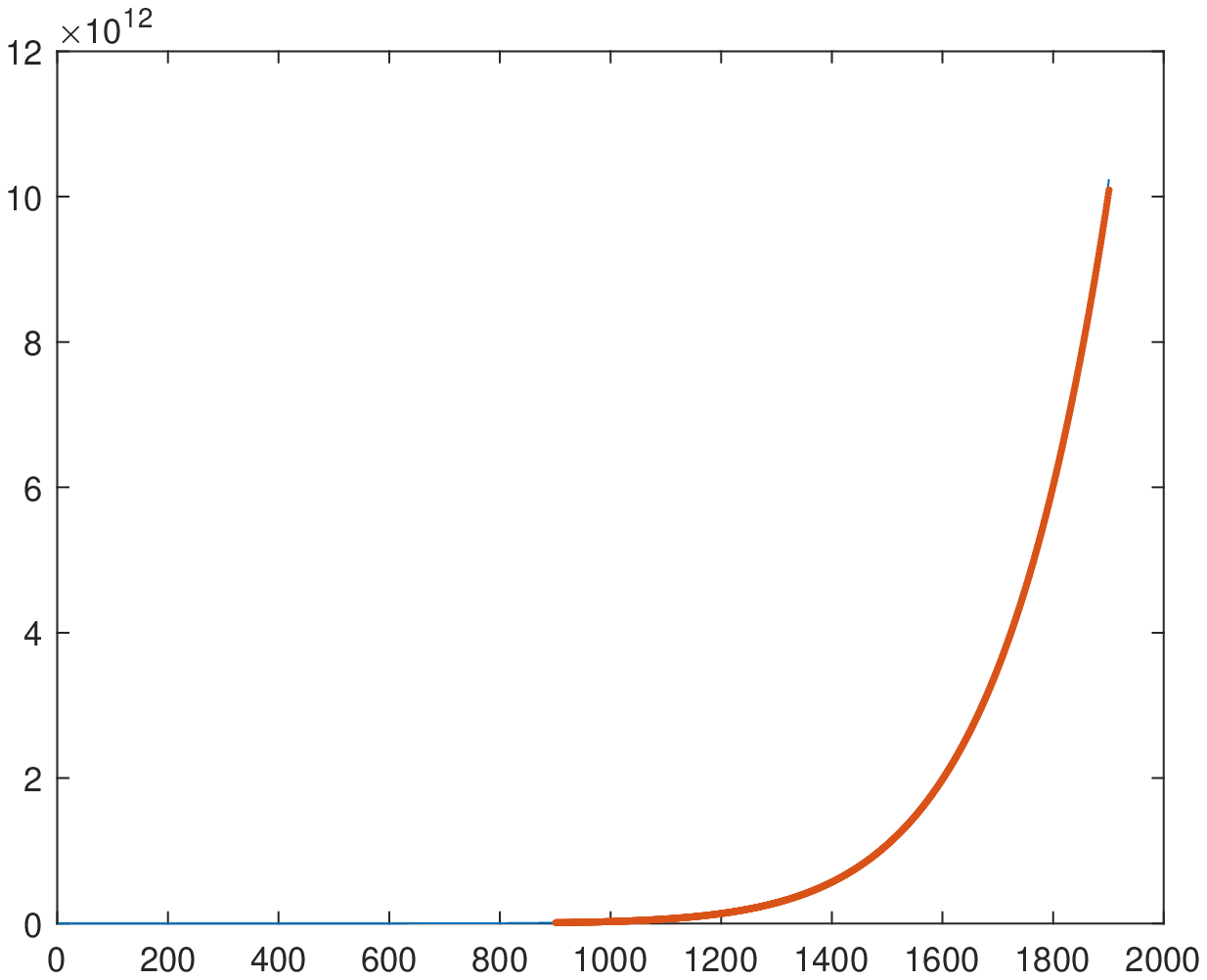}%
\label{fig3:f2}}
\subfigure[f3]{\includegraphics[width=.3\linewidth]{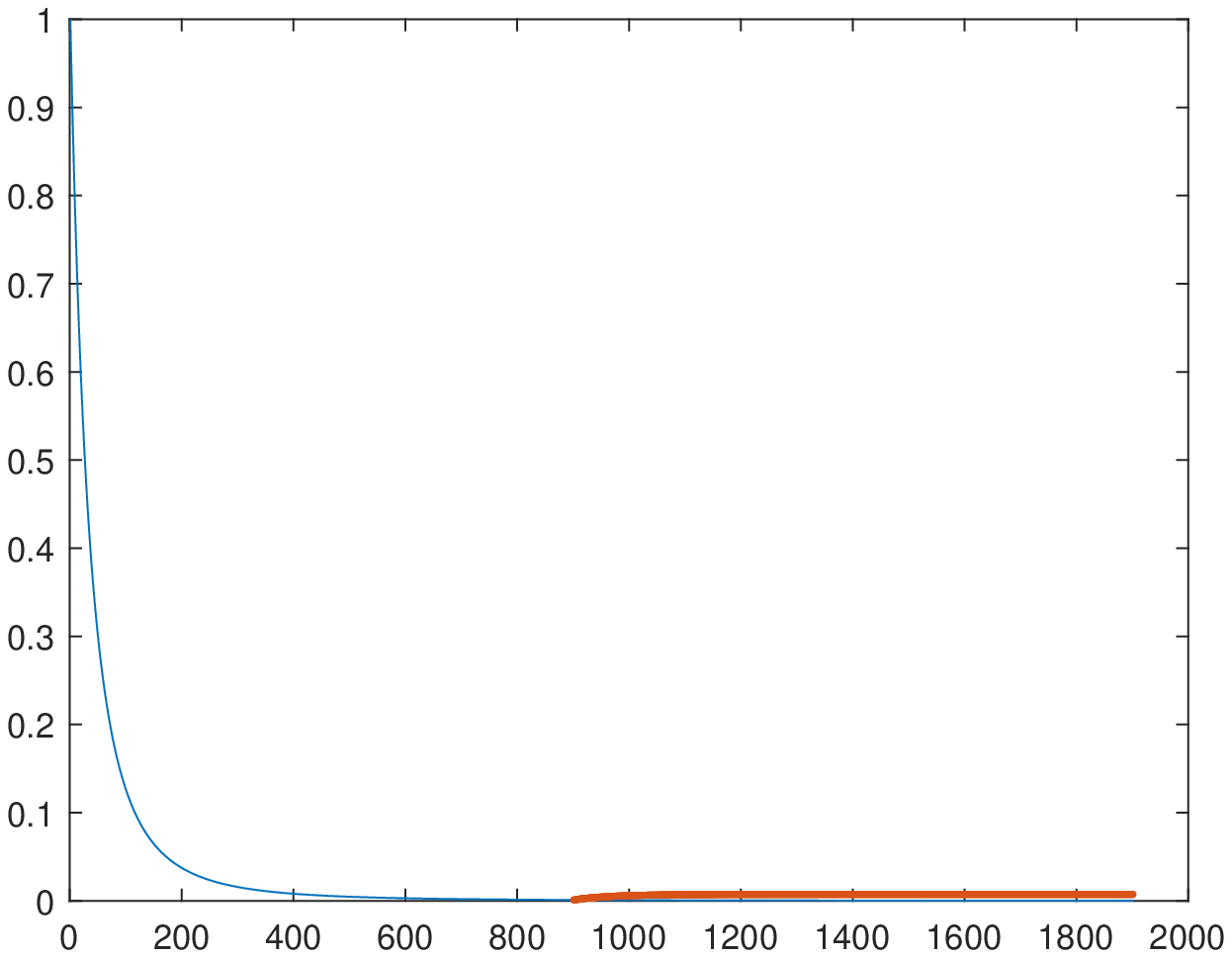}%
\label{fig3:f3}}
\hspace{.5in}
\subfigure[f4]{\includegraphics[width=.3\linewidth]{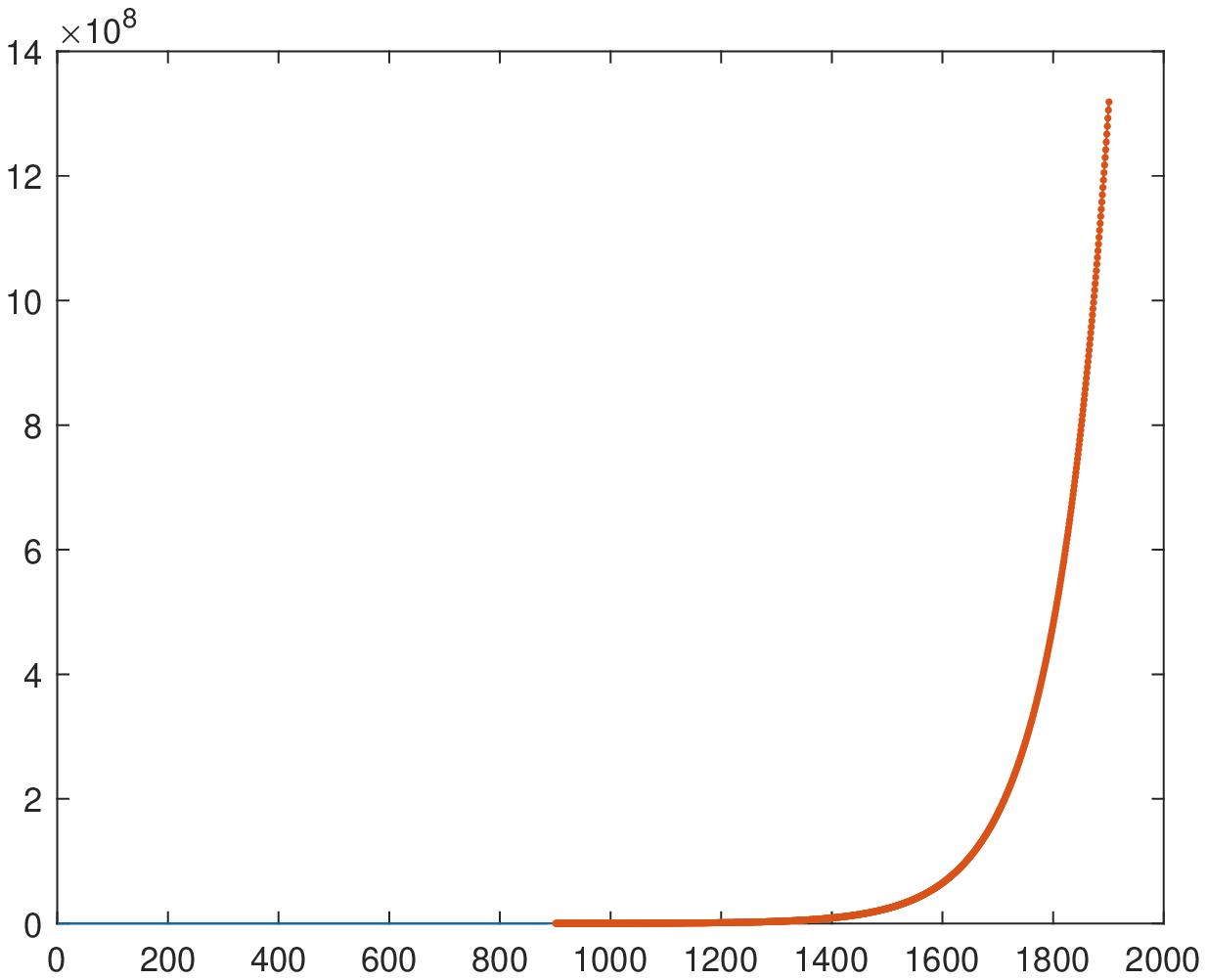}%
\label{fig3:f4}}
\subfigure[f5]{\includegraphics[width=.3\linewidth]{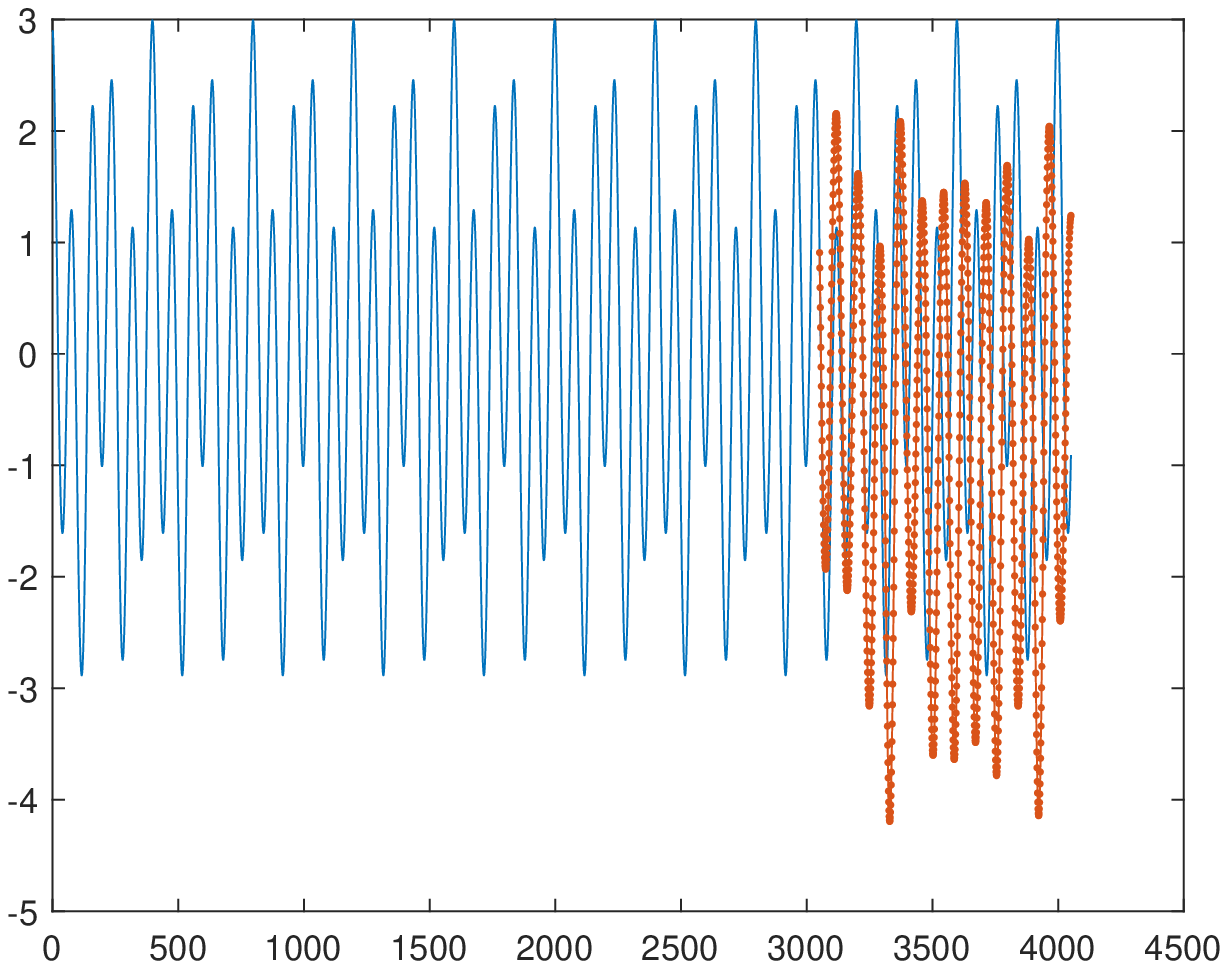}%
\label{fig3:f5}}
\subfigure[f6]{\includegraphics[width=.3\linewidth]{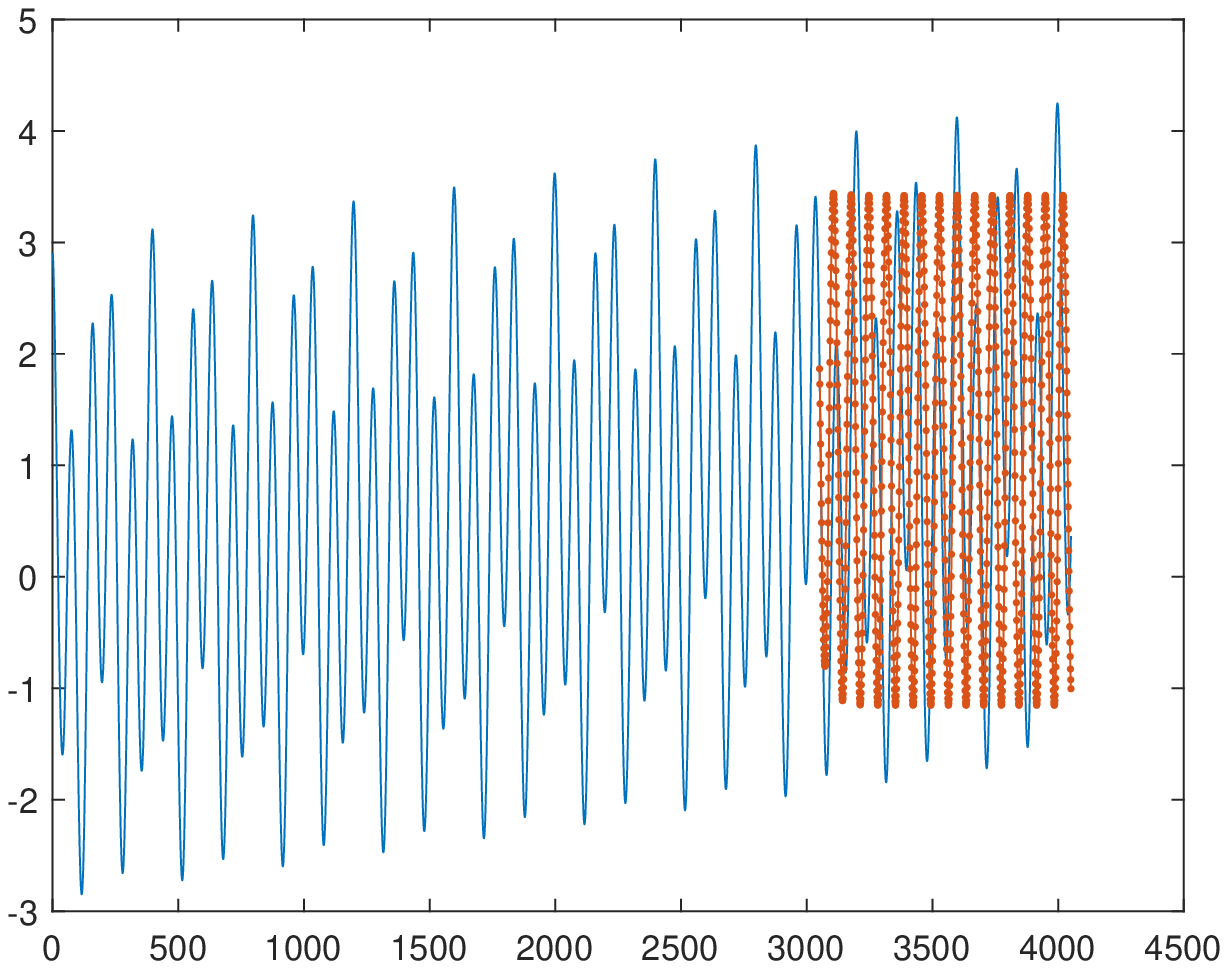}%
\label{fig3:f6}}
\caption{LSTM+ST on mathematics functions.}
\label{fig3}
\end{figure*}

\begin{figure*}[!t]
\centering
\subfigure[f1]{\includegraphics[width=.3\linewidth]{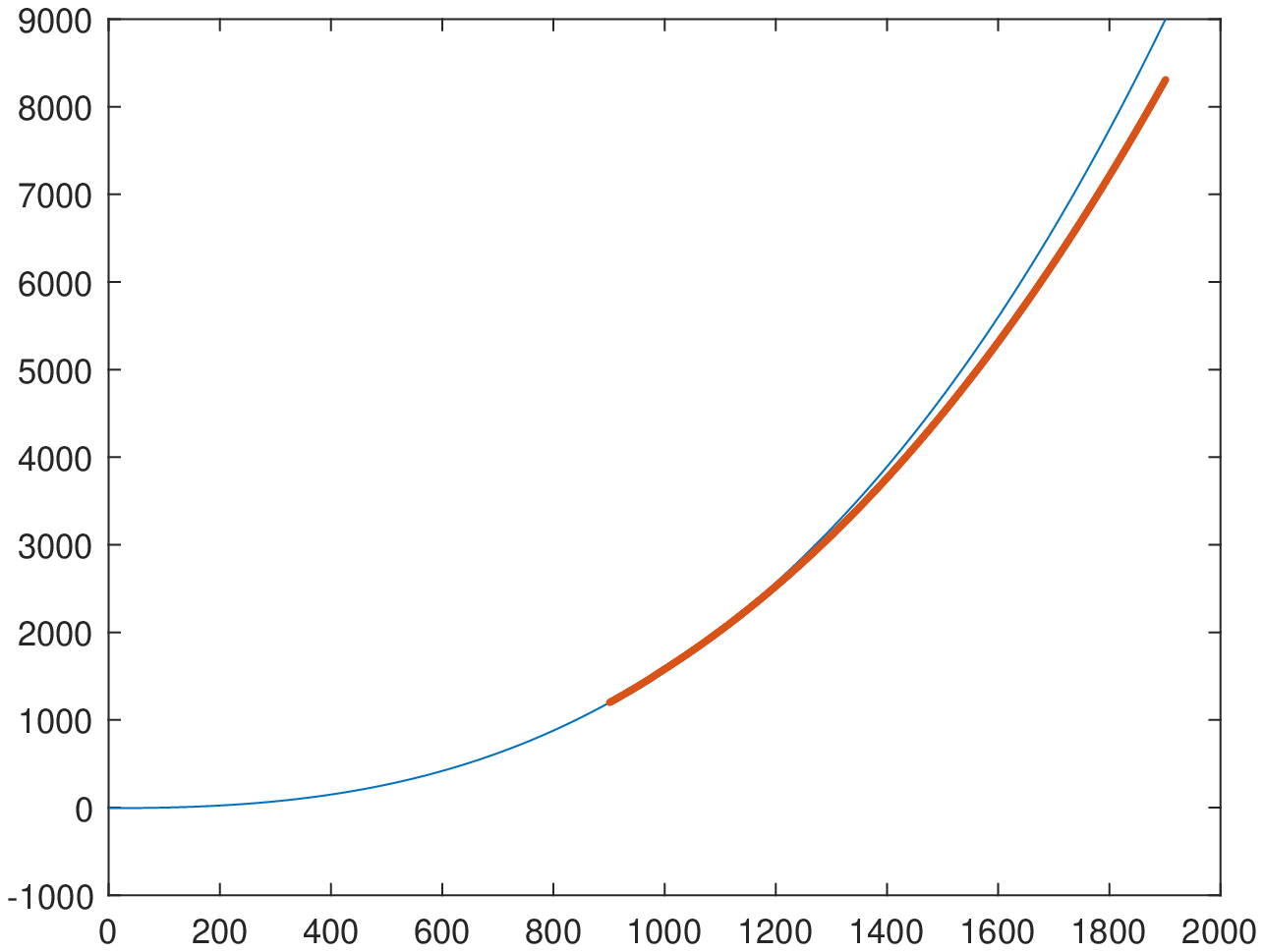}%
\label{fig4:f1}}
\subfigure[f2]{\includegraphics[width=.3\linewidth]{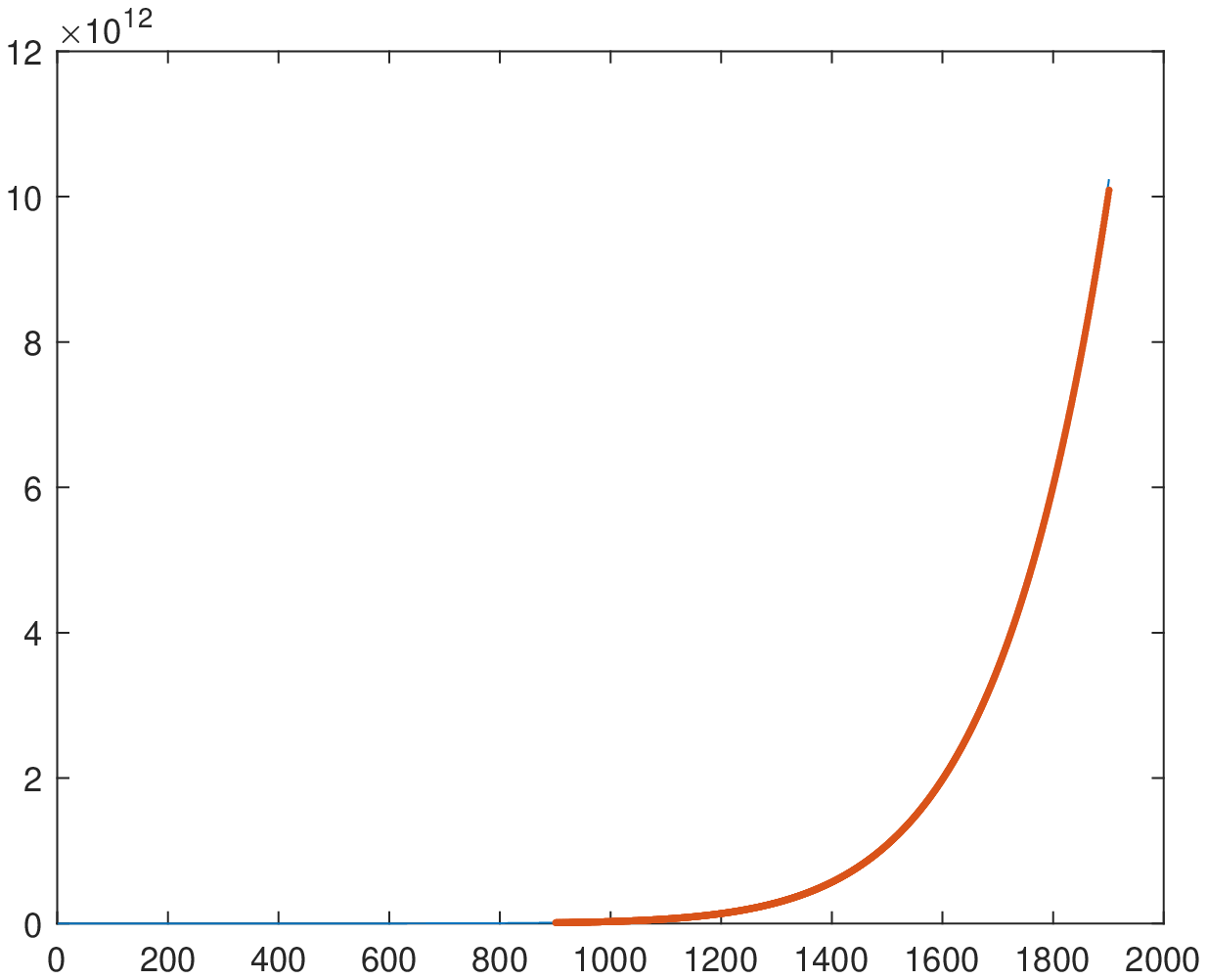}%
\label{fig4:f2}}
\subfigure[f3]{\includegraphics[width=.3\linewidth]{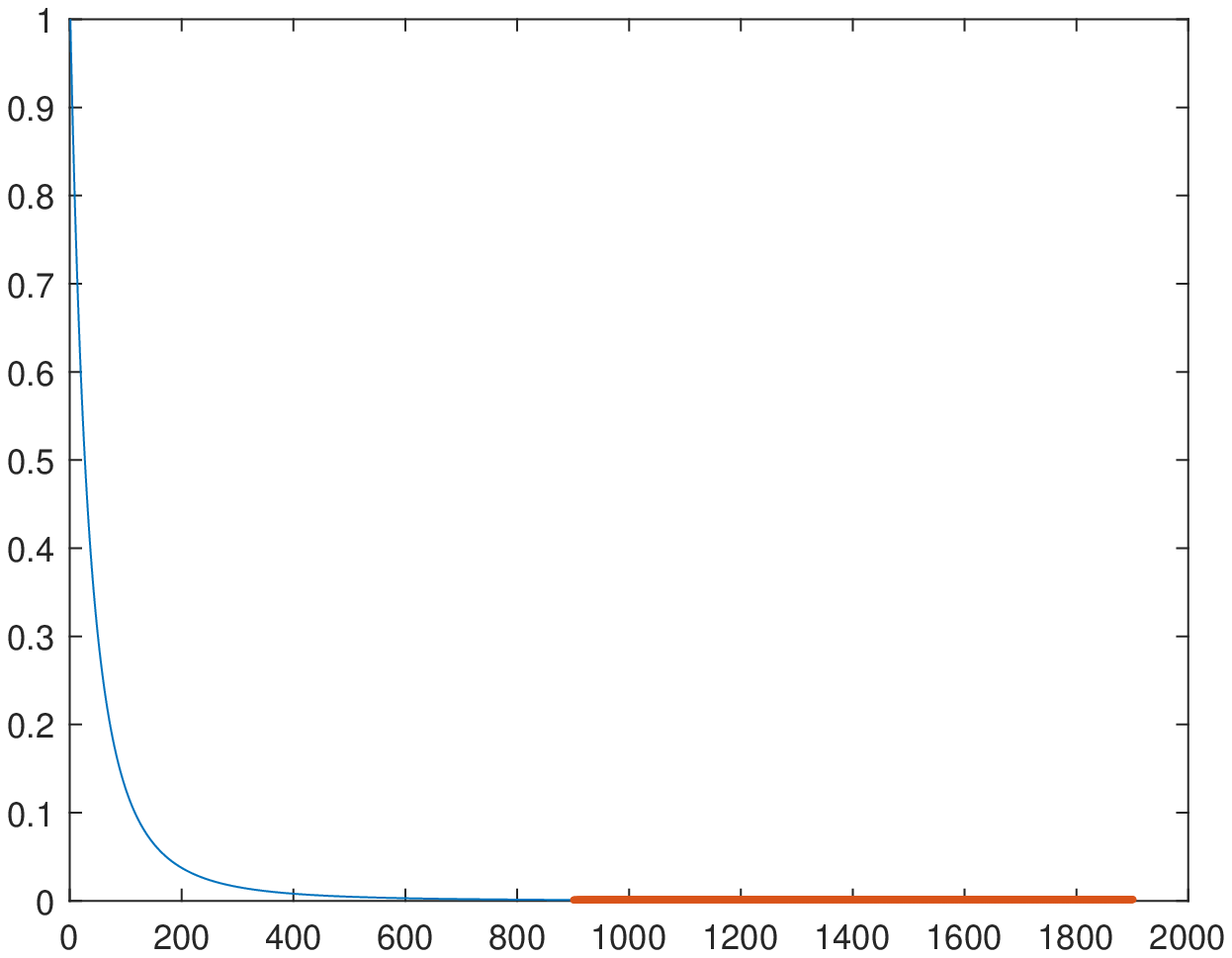}%
\label{fig4:f3}}
\hspace{.5in}
\subfigure[f4]{\includegraphics[width=.3\linewidth]{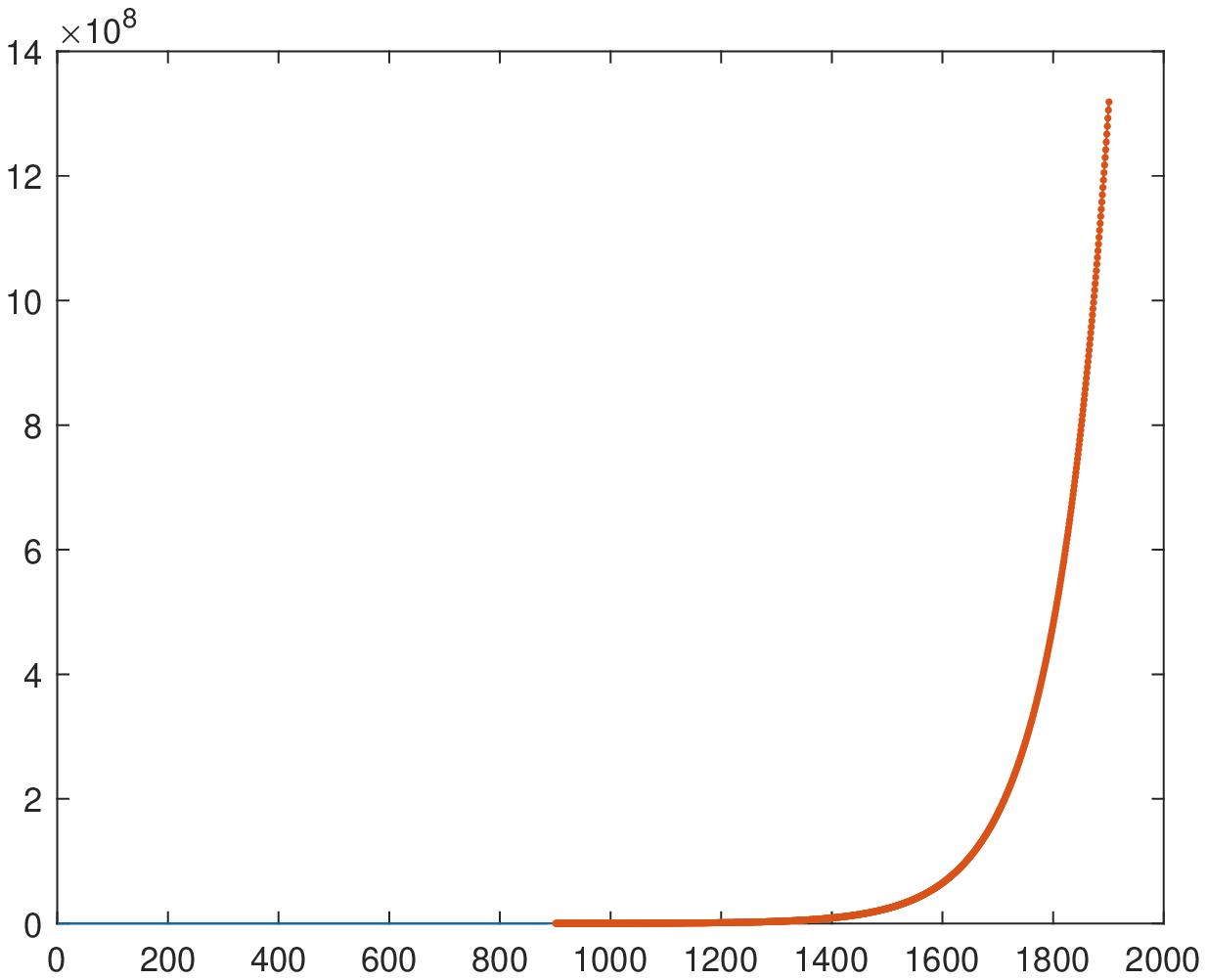}%
\label{fig4:f4}}
\subfigure[f5]{\includegraphics[width=.3\linewidth]{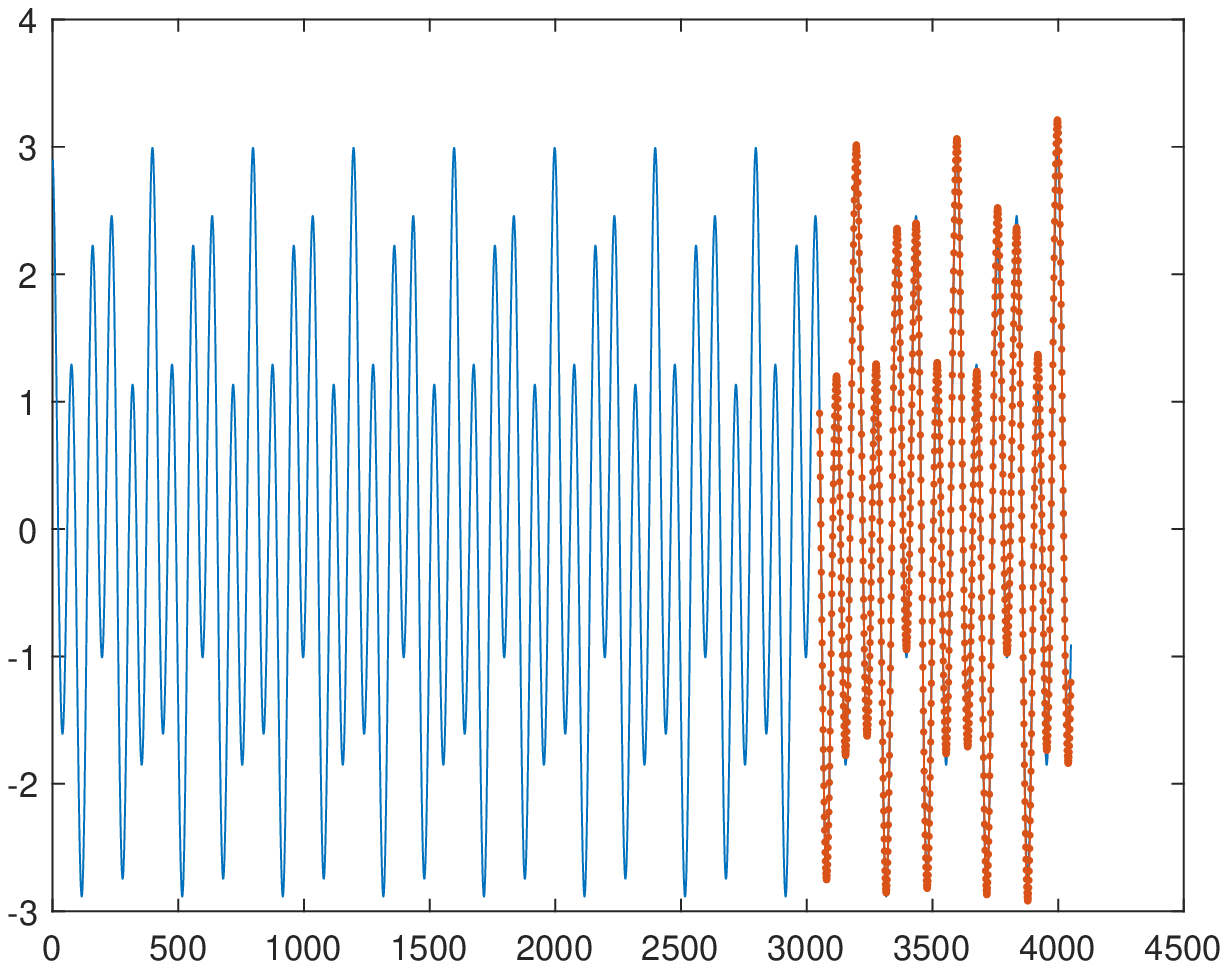}%
\label{fig4:f5}}
\subfigure[f6]{\includegraphics[width=.3\linewidth]{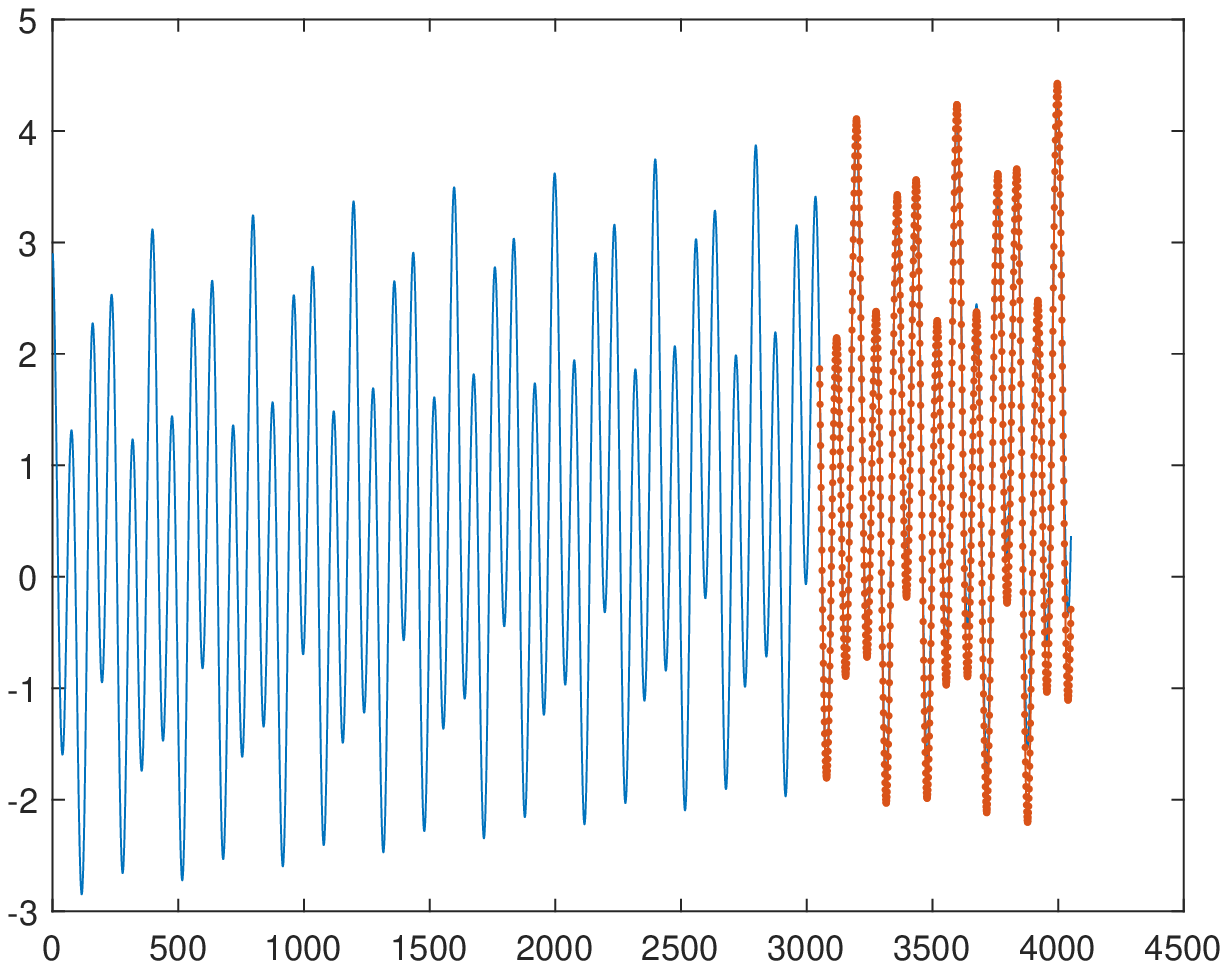}%
\label{fig4:f6}}
\caption{LRF+fitnet+ST on mathematics functions.}
\label{fig4}
\end{figure*}

As shown in Fig. \ref{fig3} and Fig. \ref{fig4}, the proposed approach can fit and generalize the mathematics functions with varying orders and periods very well. There are the middle order function f1, f5, f6, the high order functions f2, f4, the low order function f3, the periodic functions f5, f6, the non-periodic functions f1, f2, f3, f4, the stationary functions f3, f5, and the non-stationary functions f1, f2, f4, f6. The general fitting error is very small for all these conditions. The Stationary Transform also contributes to the comparing LSTM method, the LSTM+ST method also fits and generalizes the mathematics functions with varying orders and periods. It is notable that the LRF method doing much better than LSTM in f5 and f6, which has complex structure and period. It shows that the proposed method have better detail or local change fitting ability.

\begin{table*}[htb]
  \caption{The long term general fitting errors (MAE) on the 6 mathematics functions}

\begin{tabular}{lcccc}
\toprule
    & \multicolumn{2}{c} {LSTM+ST}  & \multicolumn{2}{c} {LRF+fitnet+ST} \\
    \midrule
    Function(Max) & 1-500 & 501-1000 & 1-500 & 501-1000\\
    \midrule
    F1(9000) & 67.28 &  419.37 & 55.95 & 396.85\\
    F2(1.0240e+13) & 1.42e+9 &  6.12e+10 & 1.41e+9 & 6.15e+10\\
    F3(1) & 0.0021 &   0.0025 & 0.0019  &  0.0024\\
    F4(4.8517e+08) & 120.27 &  124.66 & 120.17 &  120.29\\
    F5(3) & 1.85 &   3.28 & 0.04  &  0.05\\
    F6(10.8145) & 2.36 &    2.17 & 0.12 &  0.35\\
    \bottomrule
    \end{tabular}
    \label{tab1}
\end{table*}

As shown in table \ref{tab1}, the mean squared error of the proposed approach is usually less than the LSTM baseline approach. For the non-periodic functions f1, f2, f3, f4, the mean squared error is very low (less than $1\%$) comparing to the max value of the functions. For the periodic functions f5, f6, the mean squared error is also quite low (less than $10\%$) comparing to the max value of the function.

\begin{figure*}[!t]
\centering
\subfigure[LSTM+ST on airline]{\includegraphics[width=.45\linewidth]{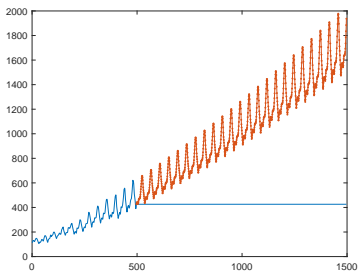}%
\label{fig5:air1}}
\subfigure[LSTM+ST on sunspot]{\includegraphics[width=.45\linewidth]{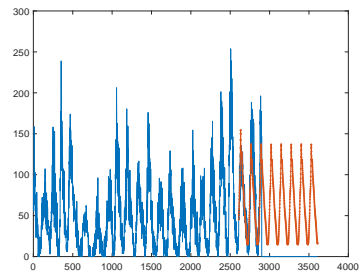}%
\label{fig5:sun1}}
\hspace{.5in}
\subfigure[LRF+fitnet+ST on airline]{\includegraphics[width=.45\linewidth]{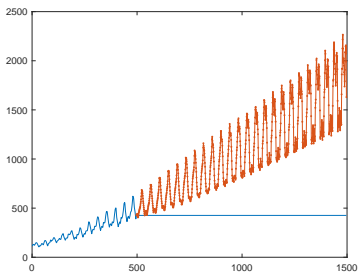}%
\label{fig5:air2}}
\subfigure[LRF+fitnet+ST on sunspot]{\includegraphics[width=.45\linewidth]{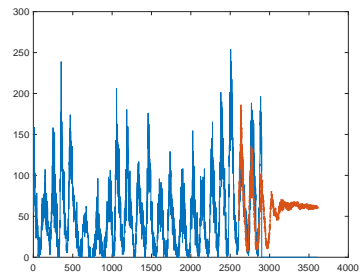}%
\label{fig5:sum2}}
\caption{Predicable function fitting on real world functions.}
\label{fig5}
\end{figure*}

As shown in Fig. \ref{fig5}, in the long term prediction test on real world functions, the proposed approach correctly generalizes the given real-world functions with different trends and periods, while the LSTM+ST method is also working well. The Stationary Transforms is working well in real world function fitting.

\begin{table*}[htb]
  \caption{The short term general fitting errors (MSE) on the 6 real world functions}

\begin{tabular}{lccccc}
\toprule
    & LSTM & Informer  & \multicolumn{3}{c} {LRF based methods} \\
    \midrule
    Function & LSTM & Informer & LSBoost & XGBoost & fitnet\\
    \midrule
    ETTm1 & 0.5440 & 0.2770 & 0.2565 & 0.2848 & \underline{\textbf{0.0289}}\\
    electricity & 0.3427 & 0.2209 &  0.7403 & 1.0524 & \underline{\textbf{0.1309}}\\
    exchangerate & \underline{\textbf{0.0369}} & 3.9276 & 0.2741 &  0.4502  &  0.0406\\
    nationalillness & 2.3452 & 4.0177 & 7.4846 & 3.3707 &  \underline{\textbf{1.2271}}\\
    traffic & 0.2097 & \underline{\textbf{0.0899}} & 1.6443 & 2.1516  &0.1080\\
    weather & 1.4267 &  1.5970 &  1.7890 & 4.1388 &  \underline{\textbf{0.5084}}\\
    \bottomrule
    \end{tabular}
    \label{tab2}
\end{table*}

As shown in table \ref{tab2}, in the short term prediction test, it has an overall test sequence of length 672 as input and a overall test sequence of length 67 as output. The mean squared error of the recent Informer Transformer network is less than the error of baseline LSTM in 3 data sets and larger than the error of LSTM in the other 3 data sets. The mean squared error of the proposed approach (LRF+LSBoost, LRF+XGBoost) is about $20\%-700\%$, which is not comparable to the result of $3\%-230\%$ of LSTM baseline approach. However, the mean squared error of $2\%-120\%$ of the proposed approach (LRF+fitnet) is much less than the result of LSTM baseline approach. In average, LRF+fitnet has the increase of prediction correction of $+56\%$ than the LSTM. This is because of that the proposed method have better capability of capturing the detail or local change of the test data. The large performance difference between the LRF+LSBoost, LRF+XGBoost and LRF+fitnet is a result of fusion global character of LRF feature. Each dimension of LRF is a global fusion of backward data, while the decision tree based methods (LSBoost, XGBoost) needs the each dimension is a distinguish individual feature.

\section{Conclusions and Discussion}\label{sec-5}
Function Fitting plays an important role in the mathematics, and it has a wide range of application in the industry, such as dynamic system modeling, time series analysis, etc... However, as far as we know, there are not many approaches that can deal with the complex data, such as stationary and non-stationary data, periodic and non-periodic data, high order and low order data. In this paper, a novel linear regression feature based generalizable function fitting technique is introduced. This approach provides a new linear regression feature transform that is served as a function analysis method. A given function is decomposed to a series linear functions layer by layer, and the regression approximate output value of these linear functions are recorded as the feature vectors of the given function. Furthermore, the one step ahead value of the given function is used as the regression targets. Therefore, a regression optimizer is used to train a regressor (fitting function). This technique is naturally capable to deal with the stationary and non-stationary functions, the periodic and non-periodic function for the middle order functions below. For the high order functions, this approach uses a stationary transform to transform the given high order function to nearly stationary functions before it is processed as the input of linear regression feature decomposition.

As we can find, the proposed approach achieves high quality results in the computational experiments. The reason of its good fitting and generalizing ability is that the LRF decomposition reserves the main characters of the given functions when it transforms the $(x,y)$ pairs to feature vectors and target vector as the input of regression optimizer. Benefit from the advance of linear and non-linear regressor, which can optimize both the training error and the general fitting error, the proposed predicable function fitting approach generates the high-quality fitting functions. Thus, it has the potential to be developed as a powerful function fitting and analysis tool. It is noticeable that the proposed approach can modelling the high-order functions, including exponential functions, by purely linear methodology. This indicates the possibility of beyond the computability gap of non-polynomial and polynomial. However, there is still room for improvement in this work. The generalizable fitting ability for functions with many details, such as pseudo random walk, is still not sufficient.




\bibliographystyle{sn-mathphys}
\bibliography{linet}


\end{document}